\documentclass{article}
\usepackage{spconf,amsmath,graphicx, multirow, tikz, makecell, verbatim}

\title{Residual-Guided Learning Representation for Self-Supervised Monocular Depth Estimation}
\name{Byeongjun Park \qquad Taekyung Kim \qquad Hyojun Go \qquad Changick Kim}
\address{School of Electrical Engineering, KAIST, Daejeon, Republic of Korea}
\begin{document}
%
\maketitle

\begin{abstract}
Photometric consistency loss is one of the representative objective functions commonly used for self-supervised monocular depth estimation. However, this loss often causes unstable depth predictions in textureless or occluded regions due to incorrect guidance. Recent self-supervised learning approaches tackle this issue by utilizing feature representations explicitly learned from auto-encoders, expecting better discriminability than the input image. Despite the use of auto-encoded features, we observe that the method does not embed features as discriminative as auto-encoded features. In this paper, we propose residual guidance loss that enables the depth estimation network to embed the discriminative feature by transferring the discriminability of auto-encoded features. We conducted experiments on the KITTI benchmark and verified our method's superiority and orthogonality on other state-of-the-art methods.
\end{abstract}


\section{Introduction}
Accurate depth estimation is essential for many computer vision applications, such as autonomous driving and robotics navigation. 
In these days of explosively increasing raw data, self-supervised monocular depth estimation has emerged as a promising approach. 
However, conventional self-supervised monocular depth estimation methods \cite{zhou2017unsupervised, yang2017unsupervised, yin2018geonet} often adopt self-supervision losses, such as a photometric consistency loss and a feature consistency loss despite the underlying side effects. 
Specifically, the photometric consistency loss tends to match nearby wrong pixels of similar appearance, while the feature consistency loss often results in the local minimum problem, causing degenerative discriminability.

A recent self-supervised depth estimation method \cite{shu2020feature} proposes a feature-metric loss that utilizes the discriminability of auto-encoded feature representations, expecting explicitly trained auto-encoder to have better capacity on encoding more discriminative information in textureless or occluded regions.
Despite the reasonable feature enhancement, we observed that the auto-encoded feature and the original depth feature of the depth estimation network derive different loss landscapes, especially in the erroneous regions, meaning that the discriminability of the depth feature is not sufficiently improved at the level of the auto-encoded feature.

In this paper, we focus on addressing the inaccurate depth predictions on the erroneous regions by sufficiently transferring the auto-encoded feature's discriminability to the depth feature. 
We propose a novel residual guidance loss with a self-supervision framework that minimizes the difference between residual depths derived by the auto-encoded feature and the depth feature.
Our method is based on the insight that harmonizing the residual depths of auto-encoded features and depth features leads to the assimilation of their loss landscapes.
We adopt the Gauss-Newton refinement approach \cite{yu2020fast} to derive residual depths from the auto-encoded feature and depth feature. We evaluate to verify the effectiveness and show that our method outperforms other state-of-the-art methods on the KITTI 2015 benchmark.

In summary, our major contributions are summarized as follows;
(a) We provide a mathematical explanation that residual depths guide the discriminability information, 
(b) Residual-guidance loss is proposed to match residual depths between features generated by auto-encoder and depth network.

\section{Proposed Method}
In this section, we first review the learning objective for training the depth and pose estimation network. Then we describe our key idea and residual guidance loss to match residual depths between auto-encoded features and depth features.
Note that we express the depth map for the target view $I_{t}$ as $D$, and the relative camera pose for each source view $I_{t'}$ with respect to the target view as $T_{t \to t'}$.

\begin{figure*}[t]
\begin{minipage}[b]{.24\textwidth}
  \centering
  \centerline{\includegraphics[width=4.3cm]{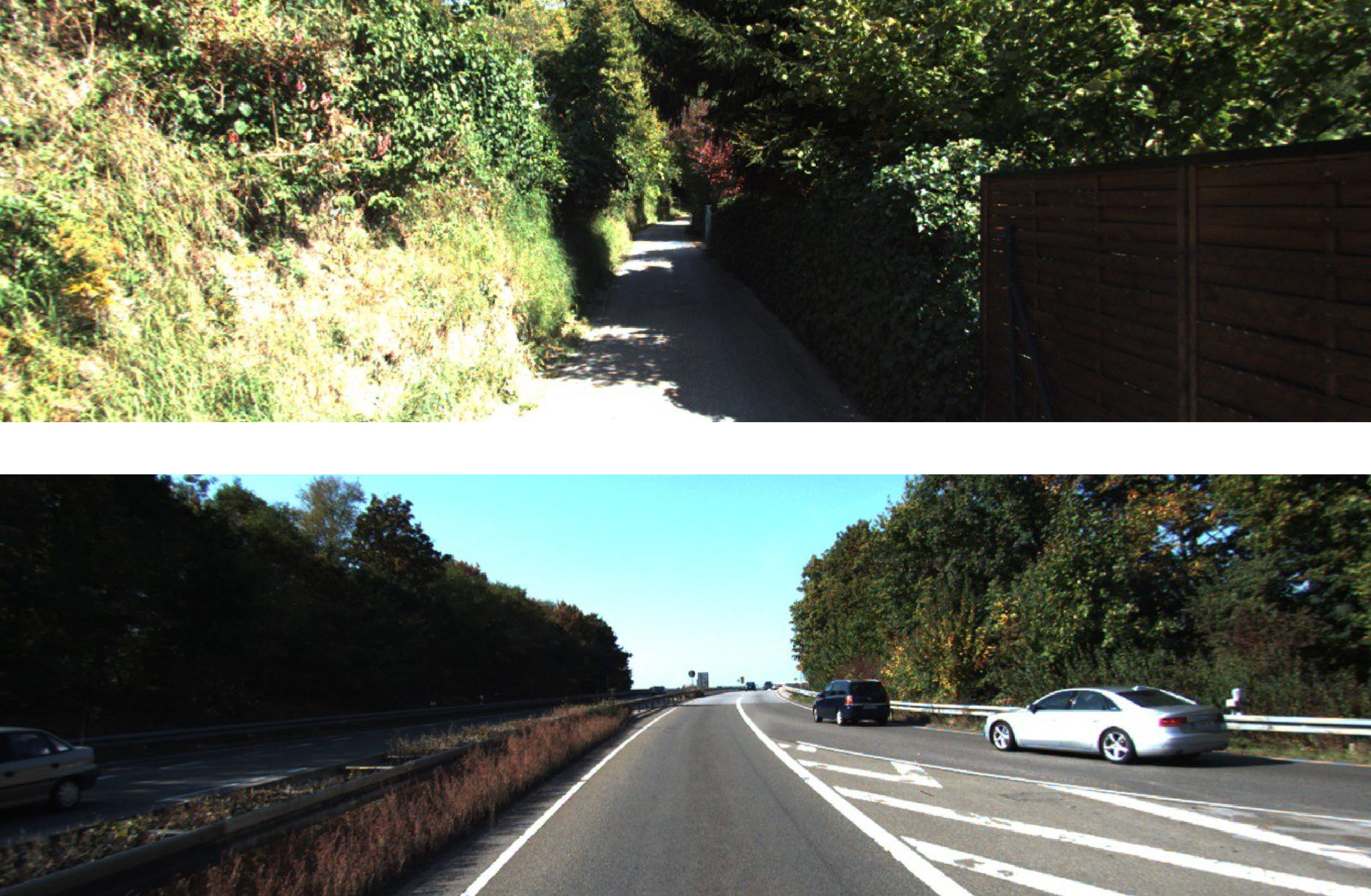}}
  \vspace{0.1cm}
  \centerline{(a) Input Images}\medskip
\end{minipage}
\hfill
\begin{minipage}[b]{.24\textwidth}
  \centering
  \centerline{\includegraphics[width=4.3cm]{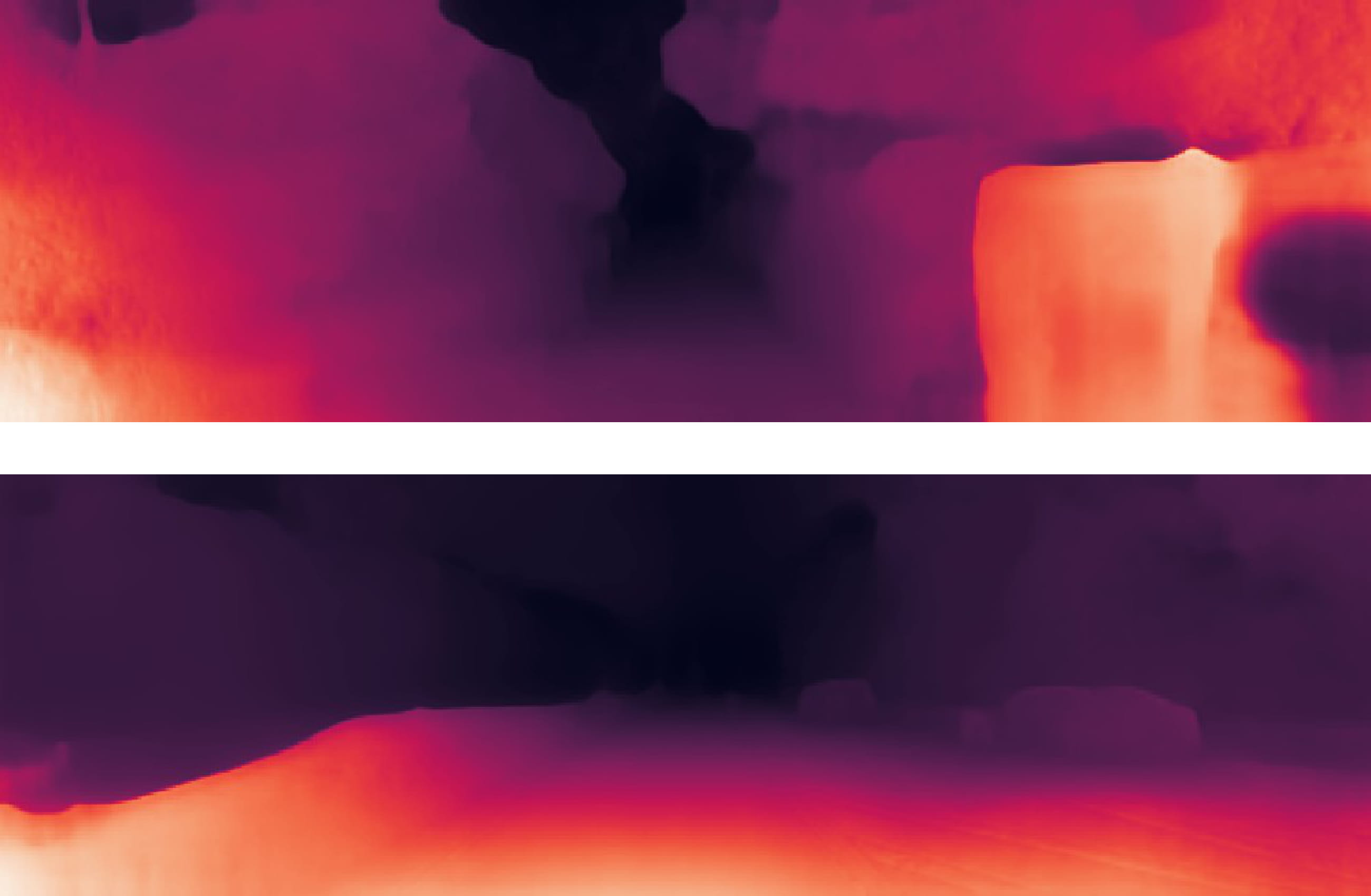}}
  \vspace{0.1cm}
  \centerline{(b) Monodepth2. \cite{godard2019digging}}\medskip
\end{minipage}
\hfill
\begin{minipage}[b]{.24\textwidth}
  \centering
  \centerline{\includegraphics[width=4.3cm]{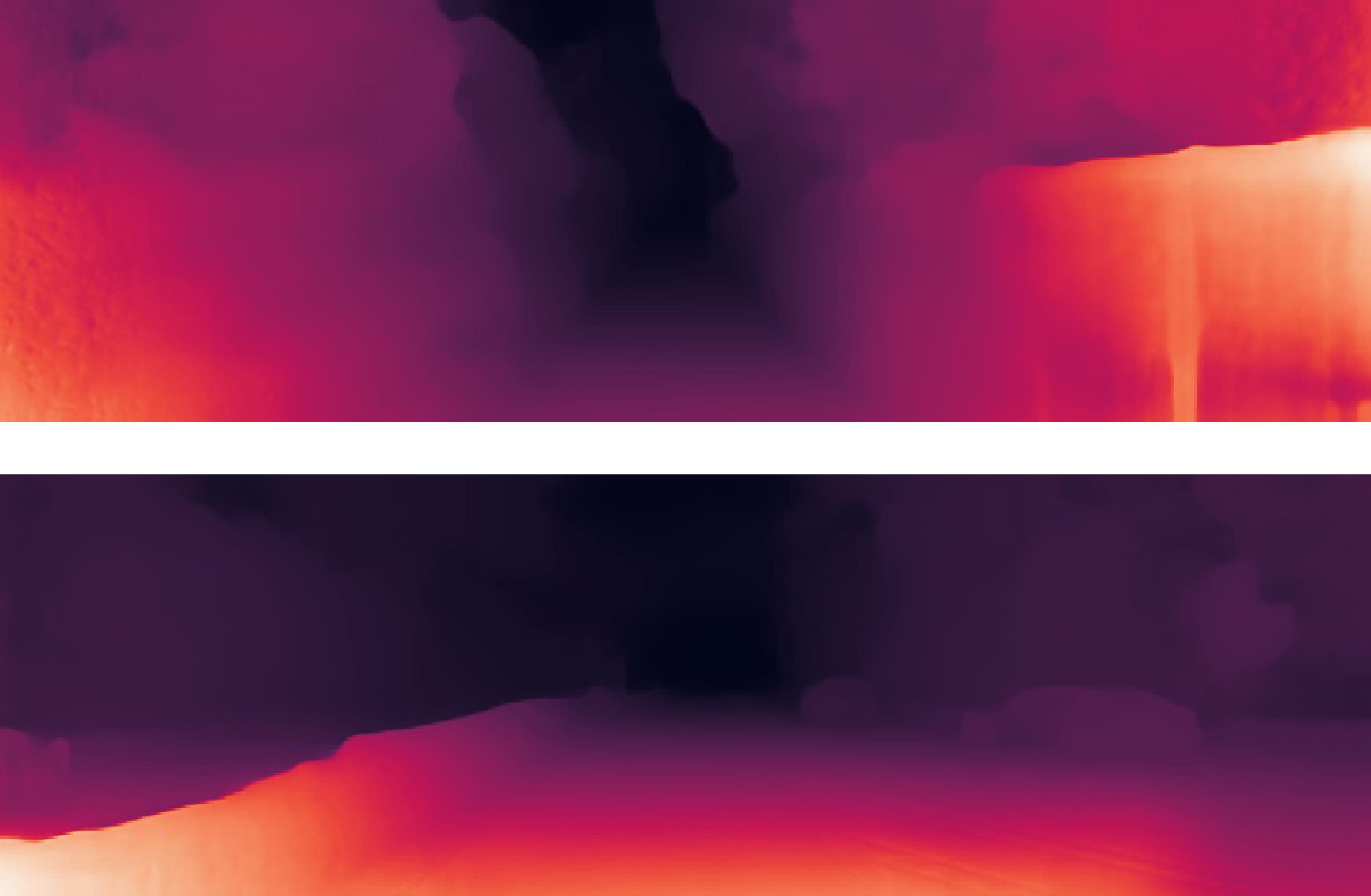}}
  \vspace{0.1cm}
  \centerline{(c) Shu et al. \cite{shu2020feature}}\medskip
\end{minipage}
\hfill
\begin{minipage}[b]{.24\textwidth}
  \centering
  \centerline{\includegraphics[width=4.3cm]{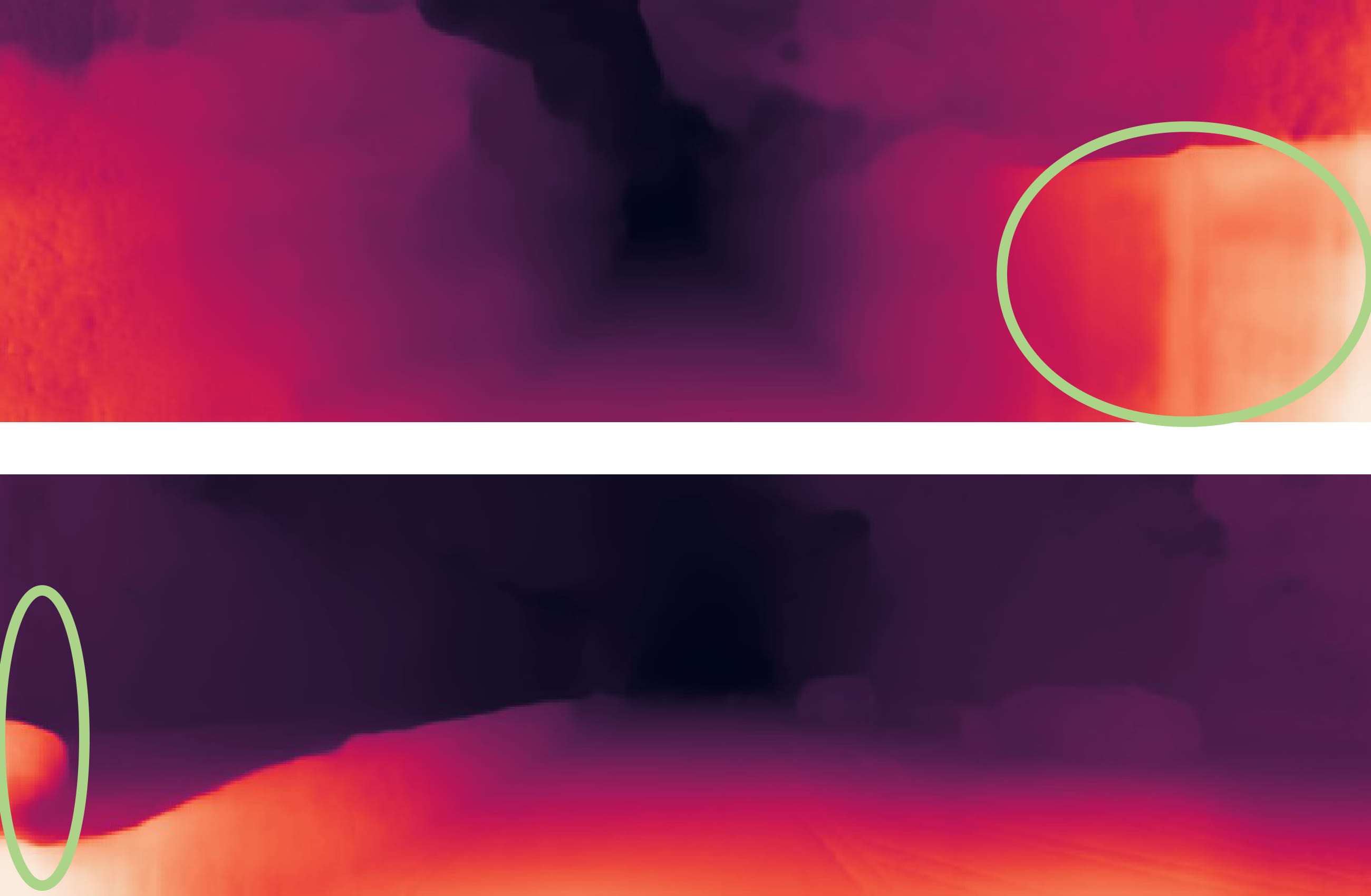}}
  \vspace{0.1cm}
  \centerline{(d) Ours}\medskip
\end{minipage}
\vspace{-0.4cm}
\caption{Example of monocular depth estimation based on self-supervision from monocular video sequences. Our self-supervised model produces high-quality depths map on out-of-view pixels and textureless regions.}
\vspace{-0.2cm}
\label{qual}
\end{figure*}

\subsection{Explicitly Constrained Feature Representation}
Our proposed architecture employs auto-encoded feature representations learned independently by an auto-encoder. Our auto-encoder are trained to reconstruct the target view $\tilde{I}_{t}$. We formulate the learning objective for training the auto-encoder based on \cite{shu2020feature}, i.e.,
\begin{equation}
L_{AE} = L_{rec} + \lambda_{dis}\, L_{dis} + \lambda_{cvt}\, L_{cvt}.
\label{AE_loss}
\end{equation}
Here, the image reconstruction loss $L_{rec}$ is an L1-loss between the target view and the reconstructed target view. We use the same discriminative loss $L_{dis}$ and convergent loss $L_{cvt}$ as \cite{shu2020feature}, expecting the auto-encoder to encode feature maps more discriminative than input images. We set $\lambda_{dis}$ to $10^{-3}$ and $\lambda_{cvt}$ to $10^{-3}$.

\subsection{Self-Supervised Training}
Our geometric self-supervision is based on \cite{shu2020feature, godard2019digging}, which consists of a photometric loss, a feature metric loss, and a smoothness term.
Similar to \cite{godard2019digging, garg2016unsupervised, godard2017unsupervised}, we formulate the photometric loss  $L_{ph}$ and the edge-aware depth smoothness loss $L_{sm}$ for self-supervision on depth prediction, i.e.,
\begin{equation}
L_{ph} = \frac{\alpha}{2} \left( {1 - SSIM_{t, t'}} \right) + \left( 1 - \alpha \right){ ||I_t - I'_{t'} ||}_1,
\end{equation}
\begin{equation}
L_{sm} = {|\partial_{x}d^{*}_{t}|e^{-|\partial_{x}I_{t}|} + |\partial_{y}d^{*}_{t}|e^{-|\partial_{y}I_{t}|}}.
\end{equation}
Here, $I'_{t'}$ is a geometrically warped source view, and we fix $\alpha = 0.85$. Also, $SSIM_{t, t'}$ is the structure similarity \cite{wang2004image} between $I'_{t'}$ and $I_{t}$, and $d^{*}_{t}=d_{t}/\bar{d_{t}}$ is the mean-normalized inverse depth from \cite{wang2018learning}.

The feature-metric loss is proposed in \cite{shu2020feature}, which allows the depth estimation network to be learned from discriminative features. We formulate the feature-metric loss as the difference between the target-view auto-encoded feature $\Phi_{t}$ and warped source-view auto-encoded feature $\Phi'_{t'}$, i.e.,
\begin{equation}
L_{fm} = || \Phi_{t} - \Phi'_{t'} ||_{1}.
\end{equation}

Finally, the overall geometric self-supervised loss $L_{G}$ is a weighted sum of photometric loss $L_{ph}$, smoothness loss $L_{sm}$, and feature-metric loss $L_{fm}$, i.e.,
\begin{equation}
L_{G} = L_{ph} + L_{fm} + \lambda_{sm}\, L_{sm},
\label{G_loss}
\end{equation}
where we set $\lambda_{sm}$ to $10^{-3}$. We also incorporate learning techniques proposed in Monodepth2 \cite{godard2019digging}, namely auto-masking, per-pixel minimum reprojection loss, and depth map upsampling, to obtain improved results. 

\subsection{Residual-Guidance Module}
\begin{figure*}[t]
  \centering
  \centerline{\includegraphics[width=1.0\linewidth]{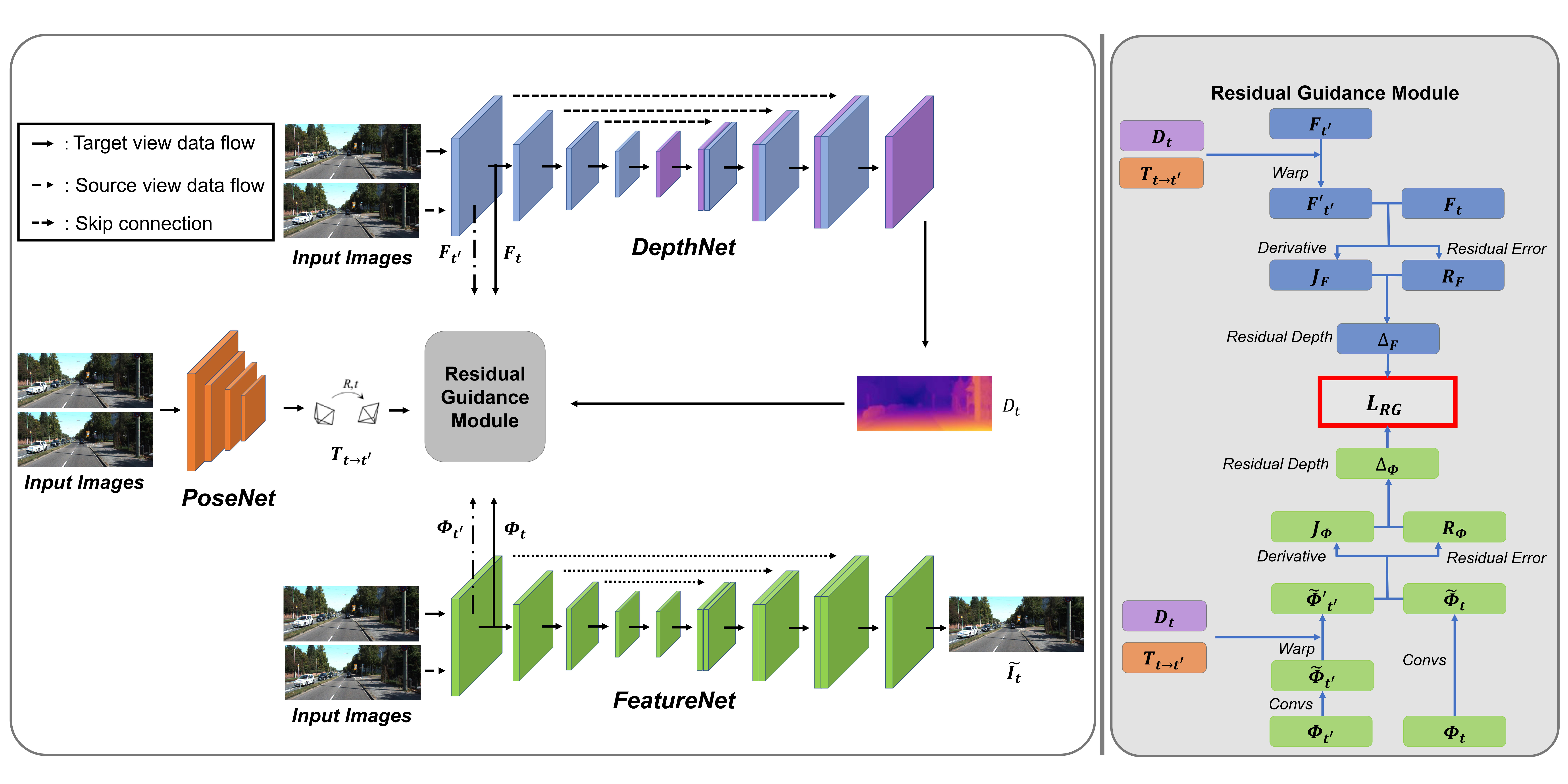}}
  \vspace{-0.4cm}
\caption{{Overview of our residual-guidance module.} Our sub-networks are illustrated in the left part. We use DepthNet, PoseNet, and FeatureNet for depth estimation, relative pose estimation, and auto-encoded feature extraction, respectively. With the estimated depth and relative camera pose, the residual guidance module computes residual guidance loss $L_{RG}$. The right part illustrates the process of formulating the residual depth according to the Gauss-Newton method.}
\label{network}
\vspace{-0.1cm}
\end{figure*}

Since the depth estimation network and the auto-encoder are trained by different supervisions, two networks derive the different loss landscapes. Thus, the discriminability of the depth feature is not as fully improved as that of the auto-encoded feature. Therefore, we focus on assimilating the loss landscapes of the features in two networks, expecting to transfer the discriminability of the auto-encoded feature to the depth feature. We match the residual depth of two differently encoded features leads to assimilate their loss landscapes. The residual depth is based on \cite{yu2020fast}, which is the increment to the current depth that minimizes the error function by the Gauss-Newton algorithm.

Let us define features $F_{t'}^{1}$, $F_{t}^{1}$, $F_{t'}^{2}$ and $F_{t}^{2}$, where $F^{1}$ and $F^{2}$ are feature representations generated by different networks for the same input image, and  $t$ and $t'$ indicate the target view and the source view, respectively. Since accurate depth estimation generally aims at minimizing photometric differences between the warped source view and the target view, we formulate photometric differences as the following error function $E_{F^{1}}$ and $E_{F^{2}}$.
\begin{equation}
      E_{F^{i}} = \sum_{p} ||F^{i}_{t'} \left( p' \right) - F^{i}_{t}\left( p \right)||_{2},
\end{equation}
for $i = \{1, 2\}$. Here, $p'$ is the warped target view pixel $p$ in source view, i.e.,
$p' = KT_{t \to t'}D\left(p\right)K^{-1}p$.
Similar to \cite{yu2020fast}, we apply the Gauss-Newton algorithm to compute residual depths $\Delta_{F^{i}}\left(p\right)$ that minimize $E_{F^{1}}$ and $E_{F^{2}}$ for the current depth and pose, i.e.,
\begin{equation}
    \Delta_{F^{i}}\left(p\right) = -\left( J_{F^{i}}^{T} \left( p \right) J_{F^{i}} \left( p \right) \right) ^{-1} J_{F^{i}}^{T} \left( p \right) R_{F^{i}} \left(p\right),
\end{equation}
where the residual error $R_{F^{i}}\left(p\right)$ and their first order derivative with respect to depth $D\left(p\right)$ is expressed as:
\begin{equation}
      R_{F^{i}}\left(p\right) = F^{i}_{t'} \left( p' \right) - F^{i}_{t}\left( p \right),
\end{equation}
\begin{equation}
    J_{F^{i}} \left( p \right) =  \frac {\partial\,  F^{i}_{t'} \left( p' \right)}{\partial p'} \cdot \frac{\partial p'}{\partial D \left(p \right)}.
\end{equation}

To show the relationship between matched residual depths and loss landscapes, we assume each feature representation has a 1-dimensional vector and omit pixel notation. Then, since $\left(J^{T}J\right)^{-1}J^{T}$ is pseudo-inverse of $J$, the residual depth is expressed as:
\begin{equation}
    \Delta_{F^{i}} = - {R_{F^{i}}} \cdot \left({\frac {\partial R_{F^{i}}}{\partial D}}\right)^{-1}.
\end{equation}

So, the equality of residual depths $(\Delta_{F^{1}}=\Delta_{F^{2}})$ formulates the following:

\begin{equation}
    {R_{F^{1}}} \cdot \left({\frac {\partial R_{F^{1}}}{\partial D}}\right)^{-1} = \, {R_{F^{2}}} \cdot \left({\frac {\partial R_{F^{2}}}{\partial D}}\right)^{-1},
\end{equation}

\begin{equation}
    || R_{F^{1}} ||  =  C \cdot || R_{F^{2}} ||,
\label{eq-motivation}
\end{equation}
where $C$ is a constant value. Eq. (\ref{eq-motivation}) implies that matching the residual depths makes the same landscape of each residual error. Through this, we assimilate the loss landscapes of the auto-encoded feature and the depth feature, expecting to transfer the discriminability of the auto-encoded feature to the depth feature.

Based on the key idea, we propose a residual guidance module. As shown in Fig. \ref{network}, we fix the auto-encoded features and apply convolutional layers with Group Normalization \cite{wu2018group} and ELU functions \cite{clevert2015fast}. We use these processed discriminative feature maps as the guidance on transferring the discriminability to depth features by matching residual depths. From this point of view, we propose the residual-guidance loss $L_{RG}$ between auto-encoded features and depth features as:
\begin{equation}
L_{RG} = \min_{t'} \log \left(1 + || {\Delta_{\Phi, t'}} - {{\Delta_{F , t'} }}||_{1} \right),
\label{RG_loss}
\end{equation}
for low-level feature maps from FeatureNet and DepthNet. Moreover, we choose the minimum residual guidance loss after calculating the residual guidance loss for all source views in the context of \cite{godard2019digging}.

\subsection{Final Training Loss}
In total, our DepthNet and PoseNet learn to minimize $L_{G}$ and $L_{RG}$, and FeatureNet learns to minimize $L_{AE}$. With Eq. (\ref{AE_loss}), Eq. (\ref{G_loss}) and Eq. (\ref{RG_loss}), the total loss is as follows:
\begin{equation}
L = L_{AE} + L_{G} + \gamma\, L_{RG}.
\end{equation}
where $\gamma$ is a weight for the residual guidance loss.
\begin{table*}[t]
\centering
\resizebox{\textwidth}{!}{
\begin{tabular}{l|cccc|ccc}
\Xhline{3\arrayrulewidth}
\multirow{2}{*}{Method}  & \multicolumn{4}{c|}{Lower is better.}   & \multicolumn{3}{c}{Higher is better.}   \\
                        &                Abs Rel    & Sq Rel    & RMSE   & RMSE log & $\delta < 1.25$  &  $\delta < 1.25^{2}$ & $\delta < 1.25^{3}$ \\ \hline \hline
Zhou et al. \cite{zhou2017unsupervised} $\dagger$                 & 0.183 & 1.595 & 6.709 & 0.270 & 0.734 & 0.902 & 0.959 \\
Yang et al. \cite{yang2017unsupervised}                  & 0.182 & 1.481 & 6.501 & 0.267 & 0.725 & 0.906 & 0.963 \\
Mahjourian et al. \cite{mahjourian2018unsupervised}     & 0.163 & 1.240 & 6.220 & 0.250 & 0.762 & 0.916 & 0.968 \\
Geonet \cite{yin2018geonet}     $\dagger$                & 0.149 & 1.060 & 5.567 & 0.226 & 0.796 & 0.935 & 0.975 \\
DF-Net \cite{zou2018df}                                & 0.150 & 1.124 & 5.507 & 0.223 & 0.806 & 0.933 & 0.973 \\
EPC++ \cite{luo2019every}                             & 0.141 & 1.029 & 5.350 & 0.216 & 0.816 & 0.941 & 0.976 \\
Monodepth2 \cite{godard2019digging}                    & 0.115 & 0.903 & 4.863 & 0.193 & 0.877 & 0.959 & \textbf{0.981} \\ 
Shu et al. \cite{shu2020feature}    $^\ast$                    & 0.113 & 0.875 & 4.806 & 0.192 & 0.877 & 0.959 & \textbf{0.981}\\ \Xhline{2\arrayrulewidth}
Ours                                                   & \textbf{0.113} & \textbf{0.867} & \textbf{4.782} & \textbf{0.191} & \textbf{0.878} & \textbf{0.960} & \textbf{0.981} \\ \Xhline{3\arrayrulewidth}
\end{tabular}%
}
\caption{Comparison of our method to existing methods on KITTI 2015 \cite{geiger2012we} using the Eigen split. The best results in each category are in bold. All methods here take images with 640 x 192 resolution and use ResNet-18 \cite{he2016deep} as an encoder. $\dagger$ indicates the latest results from Github. $^\ast$ indicates the result for a fair comparison because no published result uses a ResNet-18 encoder and 640 x 192 resolution images.}
\label{table1}
\vspace{-0.2cm}
\end{table*}

\section{Experiments}
Here, we validated that our proposed residual-guidance loss helps with textureless regions and out-of-view pixels. We evaluated the proposed method on performing self-supervised monocular depth estimation using monocular video sequences. We also compared the proposed approach with other state-of-the-art methods, which use no extra ground-truth depths or semantic segmentation labels.
\subsection{KITTI Eigen Split}
We made a fair comparison on the KITTI 2015 stereo dataset \cite{geiger2012we}, the data split of Eigen \textit{et al.} \cite{eigen2015predicting}. Moreover, detailed ablation studies were done to show the effectiveness of the residual-guidance loss. We followed \cite{zhou2017unsupervised} to remove static frames as pre-processing. This resulted in 39,810 triplets for training and 4,424 for validation. For depth evaluation, we tested our depth model on the  KITTI testing data.
\subsection{Implementation Details}
We built our depth estimation network and pose estimation network based on \cite{godard2019digging}, and FeatureNet is on \cite{shu2020feature}. We use ResNet-18 \cite{he2016deep}, with the fully-connected layer removed, as the encoder for all sub-networks, where the encoder is pre-trained on ImageNet \cite{russakovsky2015imagenet}. Our models are trained for 40 epochs using Adam \cite{kingma2014adam} optimizer, with a batch size of 8, on a single RTX 3090 GPU. The learning rate is $5 \times 10^{-5}$ for the first 20 epochs, and then dropped to $5 \times 10^{-6}$ for the remainder. The weight of residual-guidance loss $\gamma$ is $0.1$.
\subsection{Experimental Results}
Table \ref{table1} shows performances of current state-of-the-art methods for self-supervised monocular depth estimation on KITTI 2015 dataset \cite{geiger2012we}. Our method outperformed other networks for almost all metrics. On \textbf{Sq Rel} and \textbf{RMSE} that weigh on the larger error, our method handles textureless regions throughout the image. As shown in Fig. \ref{qual}, our proposed model improves the depth estimation performance on erroneous regions. In addition, our framework produces better delineation for out-of-view objects, which means the residual-guidance module improves the discriminability of depth features even in problematic areas where geometric self-supervision is not sufficiently applied.

\begin{table}[ht]
\centering
\resizebox{0.48\textwidth}{!}{
\begin{tabular}{l|cccc}
\Xhline{3\arrayrulewidth}
\multirow{2}{*}{Method}             &  \multicolumn{4}{c}{Lower is better.}     \\
                                    &   Abs Rel    & Sq Rel    & RMSE   & RMSE log  \\ \hline \hline
baseline \cite{godard2019digging} & 0.115 & 0.903 & 4.863 & 0.193  \\ 
baseline + $L_{RG}$               & 0.114 & 0.889 & 4.821 & \textbf{0.191}  \\
baseline + $L_{fm}$               & \textbf{0.113}& 0.875 & 4.806 & 0.192  \\
baseline + $L_{RG}$ + $L_{fm}$    & \textbf{0.113}& \textbf{0.867} & \textbf{4.782} & \textbf{0.191}  \\ \Xhline{3\arrayrulewidth}
\end{tabular}%
}
\caption{Ablation on losses using auto-encoded features. Our baseline model, Monodepth2 \cite{godard2019digging}, is trained without the use of auto-encoded features. The best results in each category are in bold.}
\label{table2}
\vspace{-0.2cm}
\end{table}

\subsection{Ablation Study}
To understand the effectiveness and orthogonality of our proposed residual guidance loss, we conducted an ablation study on loss functions that leverage the auto-encoded feature. As shown in Table \ref{table2}, both residual-guidance loss and feature-metric loss \cite{shu2020feature} improved performance, and the best performance was achieved when both were used. Our proposed module achieved considerable improvements in both Sq Rel and RMSE, meaning the residual guidance module can be attached to other depth estimation approaches to generate better accurate depth maps.

\section{Conclusion}
In this work, the residual-guidance loss was proposed for self-supervised learning of depth, which assimilates loss landscapes between auto-encoded features and depth features to embed discriminative depth features. We first computed residual depths from each feature map and matched residual depths between the auto-encoded feature and the depth feature. The overall framework was end-to-end trainable in a self-supervised manner, and our proposed method improved the performance of state-of-the-art depth estimation models.

\bibliographystyle{IEEE}

\bibliography{egbib}

\end{document}